\documentclass[10pt, twocolumn, letterpaper]{article}

\usepackage[accsupp]{axessibility} 

\usepackage{cvpr}              
\usepackage{graphicx}
\usepackage{amsmath}
\usepackage{amssymb}
\usepackage{booktabs}
\usepackage[numbers, sort&compress]{natbib}
\usepackage{multirow}

\usepackage[pagebackref, breaklinks, colorlinks]{hyperref}

\usepackage[capitalize]{cleveref}
\crefname{section}{Sec.}{Secs.}
\Crefname{section}{Section}{Sections}
\Crefname{table}{Table}{Tables}
\crefname{table}{Tab.}{Tabs.}

\def\cvprPaperID{*****} 
\def\confName{CVPR}
\def\confYear{2023}

\begin{document}

\title{Leveraging TCN and Transformer for effective visual-audio fusion in continuous emotion recognition}

\author{Weiwei Zhou\footnotemark[1], Jiada Lu\footnotemark[1], Zhaolong Xiong, Weifeng Wang\\
Chinatelecom Cloud\\
{\tt \small \{zhouweiwei,lujiada,xiongzl12,wangweifeng\}@chinatelecom.cn}
}

\maketitle
\renewcommand{\thefootnote}{\fnsymbol{footnote}}
\footnotetext[1]{These authors contributed equally to this work.}
\begin{abstract}
Human emotion recognition plays an important role in human-computer interaction. In this paper, we present our approach to the Valence-Arousal (VA) Estimation Challenge, Expression (Expr) Classification Challenge, and Action Unit (AU) Detection Challenge of the 5th Workshop and Competition on Affective Behavior Analysis in-the-wild (ABAW). Specifically, we propose a novel multi-modal fusion model that leverages Temporal Convolutional Networks (TCN) and Transformer to enhance the performance of continuous emotion recognition. Our model aims to effectively integrate visual and audio information for improved accuracy in recognizing emotions. Our model outperforms the baseline and ranks 3 in the Expression Classification challenge.
\end{abstract}

\section{Introduction}
\label{sec:intro}

Facial Expression Recognition (FER) can be used in a variety of applications, such as emotion recognition in videos, facial recognition for security purposes, and even in virtual reality applications. Many facial-related tasks have achieved high accuracies, such as face recognition and face attribute recognition. Despite this, the capacity to comprehend the emotions of a person is still not adequate. The subtle distinctions between emotional expressions can lead to ambiguity or uncertainty in the perception of emotions, which makes it harder to assess the emotion of a person. Therefore, the scale of most of the FER datasets are not sufficient to build a robust model.

The appearance of AffWild  and AffWild2 dataset and the corresponding challenges \cite{kollias2019expression, kollias2022abaw, kollias2022abaw2, kollias2021analysing, kollias2020analysing, kollias2021distribution, kollias2021affect, kollias2019expression, kollias2019face, kollias2019deep, zafeiriou2017aff, 2303.01498}  boost the development of affective recognition study. The Aff-Wild2 dataset contains about 600 videos with around 3M frames. The dataset is annotated with three different affect attributes: a) dimensional affect with valence and arousal; b) six basic categorical affect; c) action units of facial muscles. To facilitate the utilization of the Aff-Wild2 dataset, the ABAW5 2023 competition was organized for affective behavior analysis in the wild.

Multi-modal emotion recognition has been proven to be a more effective approach than single-modality emotion recognition, as it can utilize the complementary information between modalities to capture a more complete emotional state while being less susceptible to various noises. This improved recognition ability and generalization ability of the model can lead to more accurate and reliable results.

Considering the fact that visual and audio information contains much emotional information,  we propose to use multi-modal features for continuous facial emotion recognition and  design a network structure based on TCN and Transformer for feature fusion. Visual and audio features are first fed into their respective TCN modules, then the features are concatenated and fed into the Transformer encoder for learning, and finally, an MLP is used for prediction. Our approach can unify visual and audio features into a temporal model, designing an efficient emotion recognition network with Transformer, thereby improving the evaluation accuracy of Valence-Arousal Estimation, Action Unit Detection, and Expression Classification.

The remaining parts of the paper are presented as follows: Sec \ref{sec:RelatedWork} describe the study of facial emotion recognition and multi-modal fusion technique. Sec \ref{sec:method} describes our methodology; Sec \ref{sec:experiment} describes the experiment details and the result; Sec \ref{sec:conclusion} is the conclusion of the paper.

\section{Related Work}
\label{sec:RelatedWork}

Many previous studies were focusing on the fusion of visual and audio features for emotion recognition.
Juan et al. \cite{ortega2019emotion} presented a network that used traditional audio features and visual features extracted with a pre-trained CNN model. 
Vu et al. \cite{vu2021multitask} built a multi-task  model for valence-arousal estimation and facial expressions prediction. The authors applied the distillation knowledge architecture for training and prediction because the dataset does not include labels for all two tasks.
One of the approaches using the multi-modal mechanism for facial emotion recognition was proposed by Tzirakis et al. \cite{tzirakis2017end}, where the visual and audio features are extracted with the CNN module and are concatenated to feed into the LSTM network.  Nguyen et al. \cite{nguyen2021deep} proposed a network consisting of a two-stream auto-encoder and an LSTM to integrate visual and audio signals for emotion recognition.   Zhang et al. \cite{zhang2020m} proposed a multi-modal multi-feature approach that extracts visual features from 3D-CNN and audio features from a bidirectional recurrent neural network. Srinivas et al. \cite{parthasarathy2021detecting} propose a transformer architecture with encoder layers to integrate audio-visual features for expression tracking. Tzirakis et al. \cite{tzirakis2021end} use attention-based methods to fuse the visual and audio features.

Previous studies have proposed some useful networks on the Aff-wild2 dataset. Kuhnke et al. \cite{kuhnke2020two} combine vision and audio information in the video and construct a two-stream network for emotion recognition and achieving high performance. Yue Jin et al. \cite{jin2021multi} propose a transformer-based model to merge audio and visual feature.

Temporal Convolutional Network (TCN) was proposed by Colin Lea et al. \cite{lea2016temporal}, which hierarchically captures relationships at low-, intermediate-, and high-level time scales. Jin Fan et al. \cite{fan2021parallel} proposed a model with  a spatial-temporal attention mechanism to catch dynamic internal correlations with stacked TCN backbones to extract features from different window sizes.

The Transformer mechanism proposed by Vaswani et al. \cite{vaswani2017attention} has achieved high performance in many tasks, so many researchers exploit Transfomer for affective behavior studies. 
Zhao et al. \cite{zhao2021former} proposed a model with spatial and temporal Transformer for facial expression analysis. Jacob et al. \cite{9577264} proposed a network to learn the relationship between action units with transformer correlation module. 

Inspired by the previous work, in this paper we proposed a multi-modal fusion model with TCN and Transformer to enhance the performance of emotion recognition.

\begin{figure*}[t]
  \centering
   \includegraphics[width=1\linewidth]{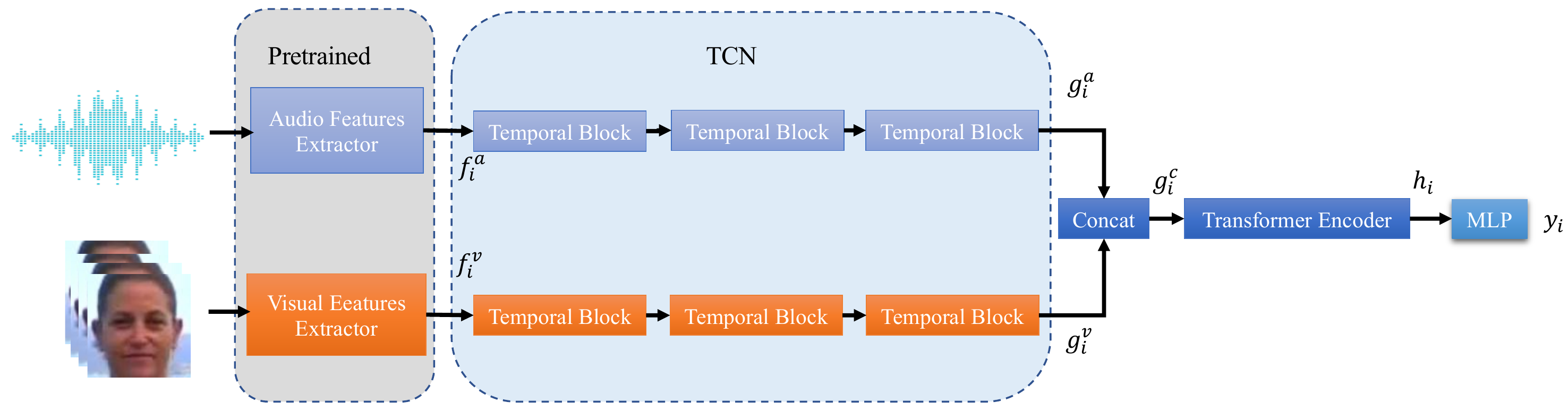}

   \caption{The architecture of our proposed model. The model consists of four components: pre-trained feature extractors for audio and visual features, TCN with three temporal blocks, Transformer encoder, and MLP for final prediction.}
   \label{fig:twocol}
\end{figure*}

\section{Methodology}
\label{sec:method}

In this section, we describe in detail our proposed method for tackling the three challenging tasks of affective behavior analysis in the wild that are addressed by the 5th ABAW Competition: Valence-Arousal Estimation, EXPR Classification, and AU Detection. We explain how we design our model architecture, data processing, and training strategy for each task and how we leverage multi-modal to improve our performance.

\subsection{Preprocessing}

We extract the audio stream from the video and preprocess it by converting it to a mono channel with a sample rate of 16, 000 Hz. This allows us to reduce the noise and complexity of the audio signal. Some of the video frames do not contain valid faces, either due to missing or not detected by the face detector. To handle this issue, we replace these frames with the closest frame that has valid face detection. This ensures that we have a consistent sequence of facial images for each video.

\subsection{Audio Features}

We use  Wav2Vec2-emotion \cite{pepino2021emotion} to extract the audio features that capture the emotional content of speech.

 Wav2Vec2-emotion is a model based on  Wav2Vec2-Large-Robust, which is pre-trained on 960 hours of LibriSpeech audio with a sampling rate of 16kHz. The model is then fine-tuned on 284 instances of MSP-Podcast data, which contains emotional speech from different speakers and scenarios. The feature vector dimension is 512, which represents a high-level representation of the acoustic signal.

To align the audio features with the video frames, we resize the features to match the length of each frame using interpolation. This ensures that we have a consistent temporal resolution for both modalities.

\subsection{Visual Features}

We extract four visual feature vectors using different models that capture various aspects of facial appearance and expression.

The first feature vector is extracted using ArcFace \cite{deng2019ArcFace} from insightface, which has been pre-trained on the Glint360K dataset \cite{an2022pfc} for face recognition. This vector encodes the identity and pose of the face with a dimension of 512.

The second feature vector is extracted using EfficientNet-b2 \cite{tan2021eff, Savchenko_2022_CVPRW}, which has been pre-trained on the VGGFace2 dataset \cite{parkhi2015deep} for face identification and fine-tuned on the AffectNet8 dataset. This vector captures the facial attributes and expressions with a dimension of 1280.

The third and fourth feature vectors are extracted using a model from DAN \cite{wen2021distract}, pre-trained on MSCeleb, and fine-tuned on RAF-DB and AffectNet8. These vectors represent the global and local features of the face with a dimension of 512 each.

\subsection{Split Videos}

Videos are first split into segments with a window size $w$ and stride $s$. Given the segment window $w$ and stride $s$, a video with $n$ frames would be split into $[n/s] + 1$ segments, where the $i$-th segment contains frames$\left\{F_{(i-1) *s+1}, \ldots, F_{(i-1) * s+w}\right\}$.

In other words, videos are cut into some overlapping chunks, each with a fixed number of frames. The purpose of doing this is to break down the video into smaller parts that are easier to process and analyze. Each chunk has some degree of overlap with the previous and next ones so that no information in the video is missed.

\subsection{Modeling}

We denote audio features as $f_i^a$ and visual features as  $f_i^v$ corresponding to the $i$-th segment.

\subsubsection{Temporal Convolutional Network}

Each feature is fed into a dedicated Temporal Convolutional Network (TCN) for temporal encoding, which can be formulated as follows: 
$$ g_i^v=\text { TCN }\left(f_i^v\right) $$ 
$$ g_i^a=\text { TCN }\left(f_i^a\right) $$ 
where $g_i^v$ denotes visual features, $g_i^a$ denotes audio features. Then, visual features and audio features are concatenated, denotes as 
$g_i^c$. 
$$
g_i^c=[g_i^v, g_i^a]
$$
This means that we use a special type of neural network that can capture the temporal patterns and dependencies of the features over time. The TCN takes the input feature vector and applies a series of convolutional layers with different kernel sizes and dilation rates to produce an output feature vector. The output feature vector has the same length as the input feature vector but contains more information about the temporal context. For example, the TCN can learn how the sound and image change over time in each segment of the video. The output feature vectors for both sound and image are then combined together by concatenating them along a dimension. This creates a new feature vector that contains both audio and visual information for each segment of the video.

\subsubsection{Temporal Encoder}

We utilize a transformer encoder to model the temporal information in the video segment as well, which can be formulated as follows: 
$$ h_i=\text { TransformerEncoder }\left(g_i^c\right). $$ 
The Transformer encoder only models the context within a single segment, thereby ignoring the dependencies between frames across segments. To account for the context of different frames, overlapping between consecutive segments can be employed, thus enabling the capture of the dependencies between frames across segments, which means $s \leq w$.

We use another type of neural network that can learn the relationships and interactions among the features within each segment. The transformer encoder takes the input feature vector that contains both audio and visual information and applies a series of self-attention layers and feed-forward layers to produce an output feature vector. The output feature vector has more semantic meaning and representation power than the input feature vector. For example, the transformer encoder can learn how different parts of the sound and image relate to each other in each segment of the video. However, the transformer encoder does not consider how different segments of the video are connected or influenced by each other. To solve this problem, we can make some segments overlap with each other so that some frames are shared by two or more segments. This way, we can capture some information about how different segments affect each other. The degree of overlap is controlled by two parameters: $s$ is the length of a segment and $w$ is the sliding window size. If $s$ is smaller than or equal to $w$, then there will be some overlap between consecutive segments.

\subsubsection{Prediction}
After the temporal encoder, the features $h_i$ are finally fed into MLP for regression, which can be formulated as follows: 
$$ y_i= \text{MLP} (h_i) $$
where $y_i $ are the predictions of $i$-th segment. For VA challenge, $y_i \in \mathbb{R}^{l \times 2}$. For EXPR challenge, $y_i \in \mathbb{R}^{l \times 8}$. For AU challenge, $y_i \in \mathbb{R}^{l \times 12}$ .

The prediction vector contains the values that we want to estimate for each segment. The MLP consists of several layers of neurons that can learn non-linear transformations of the input. The MLP can be trained to minimize the error between the prediction vector and the ground truth vector. The ground truth vector is the actual values that we want to predict for each segment. Depending on what kind of challenge we are solving, we have different types of ground truth vectors and prediction vectors. For the VA challenge, we want to predict two values: valence and arousal. Valence measures how positive or negative an emotion is. Arousal measures how active or passive an emotion is. For the EXPR challenge, we want to predict eight values: one for each basic expression (anger, disgust, fear, happiness, sadness, and surprise) plus neutral and other expressions. For the AU challenge, we want to predict twelve values: one for each action unit (AU1, AU2, AU4, AU6, AU7, AU10, AU12, 
AU15, AU23, AU24, AU25, AU26).

\subsection{Loss Functions}
VA challenge: We use the Concordance Correlation Coefficient (CCC)  between the predictions and the ground truth labels as the measure, which is defined as in Eq \ref{eq1}. It measures the correlation between two sequences $x$ and $y$ and ranges between -1 and 1, 
where -1 means perfect anti-correlation, 0 means no correlation, and 1 means perfect correlation. The loss is calculated as Eq \ref{eq2}.

\begin{equation}\label{eq1}
\begin{split}
CCC(x, y) &= \frac{2 * \operatorname{cov}(x, y)}{\sigma_x^2+\sigma_y^2+\left(\mu_x-\mu_y\right)^2} \\
\text { where } \operatorname{cov}(x, y)&=\sum\left(x-\mu_x\right) *\left(y-\mu_y\right)
\end{split}
\end{equation}

\begin{equation}\label{eq2}
\mathcal{L}_{\text {VA }}=1-CCC
\end{equation}

EXPR challenge: We use the cross-entropy loss  as the loss function, which is defined as in Eq \ref{eq3}.
\begin{equation}\label{eq3}
\mathcal{L}_{\text {EXPR }} = - \frac {1} {N}\sum_ {i} \sum_ {c=1}^My_ {ic}\log (p_ {ic})
\end{equation}

where $y_{ic}$ is a binary indicator (0 or 1) if class $c$ is the correct classification for observation $i$.
$p_{ic}$ is the predicted probability of observation $i$ being in class $c$, 
$M$ is the number of classes.
The multiclass cross entropy loss function measures how well a model predicts the true probabilities of each class for a given observation. It penalizes wrong predictions by taking the logarithm of the predicted probabilities. The lower the loss, the better the model.

AU challenge: We employ BCEWithLogitsLoss as the loss function, which integrates a sigmoid layer and binary cross-entropy,  which is defined as in Eq \ref{eq4}.

\begin{equation}\label{eq4}
\mathcal{L}_{\text {AU }} = - \frac {1} {N}\sum_ {i} [y_i\cdot log (\sigma (x_i)) + (1-y_i)\cdot log (1-\sigma (x_i))]
\end{equation}
where $N$ is the number of samples, $y_i$ is the target label for sample $i$, $x_i$ is the input logits for sample $i$, $\sigma$ is the sigmoid function
The advantage of using BCEWithLogitsLoss over BCELoss with sigmoid is that it can avoid numerical instability and improve performance.

\begin{table*}
  \centering
  \begin{tabular}{@{}lccccc@{}}
    \toprule
    Experiment & Feature & Valence  & Arousal  & F1-score  \\
    \midrule
    VA & Eff, AffectNet8, RAF-DB,  Wav2Vec2-emotion & 0.5505 & 0.6809 & -  \\
    EXPR & Eff, AffectNet8 & - & - & 0.4138  \\
    AU & Eff, AffectNet8, RAF-DB & - & - & 0.5248  \\
    
    \bottomrule
   \end{tabular}
   \caption{Performance of our method on the validation dataset of three experiments}
   \label{tab:result}
\end{table*}

\begin{table}
  \centering
  \begin{tabular}{@{}lccc@{}}
    \toprule
    Teams & Total Score & CCC-V & CCC-A  \\
    \midrule
    SituTech & \textbf{0.6414} &0.6193 & \textbf{0.6634} \\  
    Netease Fuxi &0.6372 & \textbf{0.6486} &0.6258\\
    CBCR & 0.5913 &0.5526& 0.6299 \\
    Ours & 0.5666 &0.5008 &0.6325 \\ 
    HFUT-MAC & 0.5342 &0.5234 &0.5451  \\
    HSE-NN-SberAI & 0.5048 &0.4818& 0.5279 \\
    ACCC & 0.4842 & 0.4622 &0.5062 \\
    PRL & 0.4661 & 0.5043& 0.4279  \\
    SCLAB CNU & 0.4640& 0.4578& 0.4703 \\ 
    USTC-AC & 0.2783 & 0.3245 & 0.2321 \\
    baseline & 0.201& 0.211 &0.191 \\
    \bottomrule
   \end{tabular}
   \caption{The overall test results on VA challenge. The bold fonts indicate the best results.}
   \label{tab:va_result}
\end{table}

\begin{table}
  \centering
  \begin{tabular}{@{}lccc@{}}
    \toprule
    Teams & F1  \\
    \midrule
    Netease Fuxi & \textbf{0.4121} \\
    SituTech &0.4072 \\
    Ours &0.3532 \\
    HFUT-MAC &0.3337 \\
    HSE-NN-SberAI &0.3292 \\
    AlphaAff &0.3218 \\
    USTC-IAT-United &0.3075 \\
    SSSIHL DMACS &0.3047 \\
    SCLAB CNU &0.2949 \\
    Wall Lab &0.2913 \\
    ACCC &0.2846 \\
    RT IAI &0.2834 \\
    DGU-IPL &0.2278 \\
    baseline &0.2050 \\
    \bottomrule
   \end{tabular}
   \caption{The overall test results on EXPR challenge. The bold fonts indicate the best results.}
   \label{tab:expr_result}
\end{table}

\begin{table}
  \centering
  \begin{tabular}{@{}lccc@{}}
    \toprule
    Teams & F1  \\
    \midrule
    Netease Fuxi & \textbf{0.5549} \\
    SituTech & 0.5422 \\
    USTC-IAT-United & 0.5144 \\
    SZFaceU & 0.5128 \\
    PRL & 0.5101 \\
    Ours & 0.4887 \\
    HSE-NN-SberAI & 0.4878 \\
    USTC-AC & 0.4811 \\
    HFUT-MAC & 0.4752 \\
    SCLAB CNU & 0.4563 \\
    USC IHP & 0.4292 \\
    ACCC & 0.3776 \\
    baseline & 0.365 \\
    \bottomrule
   \end{tabular}
   \caption{The overall test results on AU  challenge. The bold fonts indicate the best results.}
   \label{tab:au_result}
\end{table}

\begin{table*}
  \centering
  \begin{tabular}{@{}lcccccccc@{}}
    \toprule
    Task & Evaluation Metric & Partition & Method & Fold 0 & Fold 1 & Fold 2 & Fold 3 & Fold 4 \\
    \midrule
    \multirow{4}*{Valence} & \multirow{8}*{CCC} & \multirow{2}*{Validation} & Ours & 0.5505 & 0.6455 & 0.5889 & 0.5394 & 0.5406 \\
    ~                                 &                ~                &  ~ & Baseline & 0.24     & - & - & - & -\\
     \cline{4-9}
    ~ & ~ & \multirow{2}*{Test} & Ours & 0.5504 & 0.4979 & \textbf{0.5008} & 0.4979& 0.4875 \\
    ~                                 &                ~                &  ~ & Baseline & 0.211     & - & - & - & -\\
    \cline{3-9}
    \multirow{4}*{Arousal} & ~ & \multirow{2}*{Validation} & Ours& 0.6809 & 0.6259 & 0.6539 & 0.6468 & 0.6591 \\
    ~                                 &                ~                &  ~ & Baseline & 0.20     & - & - & - & -\\
     \cline{4-9}
     ~ & ~ & \multirow{2}*{Test} & Ours & 0.5805 & 0.5396 & \textbf{0.6325} & 0.5037 & 0.5569 \\
    ~                                 &                ~                &  ~ &Baseline &  0.191     & - & - & - & -\\

    \midrule
     \multirow{4}*{EXPR} & \multirow{4}*{ F1-score} & \multirow{2}*{Validation} & Ours & 0.4138 & 0.4350 & 0.3614 & 0.3959 & 0.4234\\
         ~ & ~ & ~ &Baseline &  0.23 & - & - & - & -\\
          \cline{4-9}
      ~ &~ & \multirow{2}*{Test} & Ours & 0.3406 & 0.2979 & \textbf{0.3532} & 0.3293 & 0.3427\\
         ~ & ~ & ~ & Baseline & 0.2050 & - & - & - & -\\

    \midrule
    \multirow{4}*{AU} & \multirow{4}*{F1-score} &\multirow{2}*{Validation} & Ours &  0.5248 & 0.5524 & 0.5000 & 0.5060 & 0.5393\\
       ~ & ~ & ~ & Baseline & 0.39 & -& - & - & -\\
        \cline{4-9}
       ~ & ~ &\multirow{2}*{Test} &  Ours & 0.4735 & 0.4822 & \textbf{0.4887} &0.4818 & 0.4720\\
       ~ & ~ & ~ & Baseline & 0.365 & -& - & - & -\\

    \bottomrule
   \end{tabular}
   \caption{Results for the five folds of three tasks}
   \label{tab:va_fold}
\end{table*}

\section{Experiments and Results}
\label{sec:experiment}

\begin{table*}
  \centering
  \begin{tabular}{@{}llcc@{}}
    \toprule
    Visual Features & Audio Features & Valence & Arousal \\
    \midrule
    ArcFace &  None & 0.5013 & 0.6054 \\
    AffectNet8 &  None & 0.5392 & 0.6629 \\
    RAF-DB &  None & 0.5109 & 0.6579 \\
    Eff & None & 0.5208 & 0.6467 \\
    Eff, ArcFace &  None & 0.5216 & 0.6519 \\
    Eff, ArcFace, AffectNet8 &  None & 0.5345 & 0.6532 \\
    Eff, ArcFace, AffectNet8, RAF-DB &  None & 0.5429 & 0.6613 \\
    Eff, ArcFace, AffectNet8, RAF-DB &  Wav2Vec2-emotion & \textbf{0.5505} & \textbf{0.6809} \\

    \bottomrule
  \end{tabular}
  \caption{Ablation study of features on the validation dataset of VA experiment.}
  \label{tab:va_validation}
\end{table*}

\begin{table}
  \centering
  \begin{tabular}{@{}llcc@{}}
    \toprule
    Visual Features & Audio Features & F1-score  \\
    \midrule
    ArcFace &  None & 0.3512 \\
    AffectNet8 &  None &  0.3937 \\
    RAF-DB &  None &  0.3928 \\
    Eff & None & 0.4018 \\
    Eff, ArcFace &  None & 0.4015 \\
    Eff, AffectNet8 &  None & \textbf{0.4138} \\
    Eff, RAF-DB &  None & 0.4012 \\
    Eff, AffectNet8, ArcFace &  None & 0.4093 \\
    Eff, AffectNet8, RAF-DB &  None & 0.4087 \\
    Eff, AffectNet8 &  Wav2Vec2-emotion & 0.4028 \\
    \bottomrule
  \end{tabular}
  \caption{Ablation study of features on the validation dataset of EXPR experiment.}
  \label{tab:expr_validation}
\end{table}

\subsection{Experiments Settings}
All models are trained on an Nvidia GeForce GTX 3090 GPU which has 24GB of memory. We use AdamW optimizer and cosine learning rate schedule with the first epoch warmup. The learning rate is $3e-5$, the weight decay is $1e-5$, the dropout prob is 0.3, and the batch size is 32.

For VA Challenge, we use Wav2Vec2-emotion, Eff, RAF-DB, and AffectNet8 as the input features.

For EXPR Challenge, we use two types of input features: Eff and AffectNet8 as described above.

For AU Challenge, we use three types of input features: Eff, RAF-DB, and AffectNet8 as described above.

For all three challenges, we split videos using a segment window $w=300$ and a stride $s=200$. This means we divide each video into segments of 300 frames with an overlap of 100 frames between consecutive segments. This helps us capture the temporal dynamics of facial expressions and emotions.

\begin{table}
  \centering
  \begin{tabular}{@{}llcc@{}}
    \toprule
    Visual Features & Audio Features & F1-score \\
    \midrule
    ArcFace &  None & 0.4598 \\
    AffectNet8 &  None &  0.4894 \\
    RAF-DB &  None &  0.4915 \\
    Eff & None & 0.5118 \\
    Eff, ArcFace &  None & 0.5042 \\
    Eff, AffectNet8 &  None & 0.5215 \\
    Eff, RAF-DB &  None & 0.5155 \\
    Eff, AffectNet8, ArcFace &  None & 0.5109 \\
    Eff, AffectNet8, RAF-DB &  None & \textbf{0.5248} \\
    Eff, AffectNet8, RAF-DB &  Wav2Vec2-emotion & 0.5134 \\

    \bottomrule
  \end{tabular}
  \caption{Ablation study of features on the validation dataset of AU experiment.}
  \label{tab:au_validation}
\end{table}

\subsection{Overall Results}
Table \ref{tab:result} displays the experimental results of our proposed method on the validation set of the VA, EXPR, and AU Challenge, where the Concordance Correlation Coefficient (CCC) is utilized as the evaluation metric for both valence and arousal prediction, and F1-score is used to evaluate the result of EXPR and AU challenge. As demonstrated in the table, our proposed method outperforms the baseline significantly. These results show that our proposed approach using TCN and Transformer-based model effectively integrates visual and audio information for improved accuracy in recognizing emotions  on this dataset.

Table \ref{tab:va_result}, Table \ref{tab:expr_result}, and Table \ref{tab:au_result} display the overall test results on the three challenges. Notably, Netease Fuxi and SituTech achieved the first and second highest scores in all three challenges, surpassing other teams significantly, indicating their exceptional performance in these challenges. Our team ranks fourth in the VA challenge, third in the EXPR challenge, and sixth in the AU challenge. Our team's performance demonstrates our competitive standing in the challenges, with notable achievements in the VA, EXPR, and AU challenges.

\subsection{Ablation Study}

In this section, we perform several ablation studies on these three experiments to compare the contribution of different features. From Table \ref{tab:va_validation}, it can be seen that almost every feature contributes to the VA prediction task, and the combination of 4 visual features: Eff, ArcFace, AffectNet8, RAF-DB, and the audio features:  Wav2Vec2-emotion reach the highest CCC score on VA experiment. Table \ref{tab:expr_validation} shows that the use of Eff and AffectNet8 can reach the highest F1-score in the EXPR experiments. Table \ref{tab:au_validation} shows that Eff, AffectNet8, and RAF-DB  can reach the highest F1-score in the EXPR and AU experiments. The cross-validation result of the VA, EXPR, and AU experiments are reported in Table \ref{tab:va_fold}. Fold 0 is exactly the original data from the ABAW dataset.

\section{Conclusion}
\label{sec:conclusion}

Our proposed approach utilizes a combination of a Temporal Convolutional Network (TCN) and a Transformer-based model to integrate visual and audio information for improved accuracy in recognizing emotions. The TCN captures relationships at low-, intermediate-, and high-level time scales, while the Transformer mechanism merges audio and visual features. We conducted our experiment on the Aff-Wild2 dataset, which is a widely used benchmark dataset for emotion recognition. Our results show that our method significantly outperforms the baseline. Finally, our team ranks fourth in the VA challenge, third in the EXPR challenge, and sixth in the AU challenge.

{\small
\bibliographystyle{ieee_fullname}
\bibliography{egbib}
}

\end{document}